%% file: main.tex
\documentclass[10pt,twocolumn,letterpaper]{article}

\usepackage{cvpr}              
\input{preamble}
\definecolor{cvprblue}{rgb}{0.21,0.49,0.74}
\usepackage[pagebackref,breaklinks,colorlinks,allcolors=cvprblue]{hyperref}

\def\confName{CVPR}
\def\confYear{2026}
\definecolor{mypink}{RGB}{239,43,159}
\title{VisionDirector: Vision-Language Guided Closed-Loop Refinement for Generative Image Synthesis}

\usepackage{pifont} 

\usepackage{siunitx}
\usepackage[table]{xcolor}

\definecolor{VDred}{HTML}{800020}
\newcommand{\cmark}{\textbf{\ding{51}}}
\newcommand{\xmark}{\textbf{\ding{55}}}

\author{
Meng Chu$^{1}$ \quad
Senqiao Yang$^{2}$ \quad
Haoxuan Che$^{3*\#}$ \quad
Suiyun Zhang$^{3}$ \quad
Xichen Zhang$^{1}$ \quad
Shaozuo Yu$^{2}$ \\
Haokun Gui$^{1}$ \quad
Zhefan Rao$^{1}$ \quad
Dandan Tu$^{3}$ \quad
Rui Liu$^{3}$\footnotemark[1] \quad
Jiaya Jia$^{1}$ \\
\\
$^{1}$The Hong Kong University of Science and Technology \\
$^{2}$The Chinese University of Hong Kong \\
$^{3}$Huawei Research \\
}

\newcommand{\mypara}[1]{\smallskip\noindent\textbf{#1}}

\begin{document}

\twocolumn[{%
\renewcommand\twocolumn[1][]{#1}%
\maketitle
\vspace{-12mm}
\begin{center}
    \large{\color{mypink} Project: \url{https://visiondirector.github.io/}}
\end{center}
\vspace{-0.5cm}
\begin{center}
\captionsetup{type=figure}
\centering
\includegraphics[width=\linewidth]{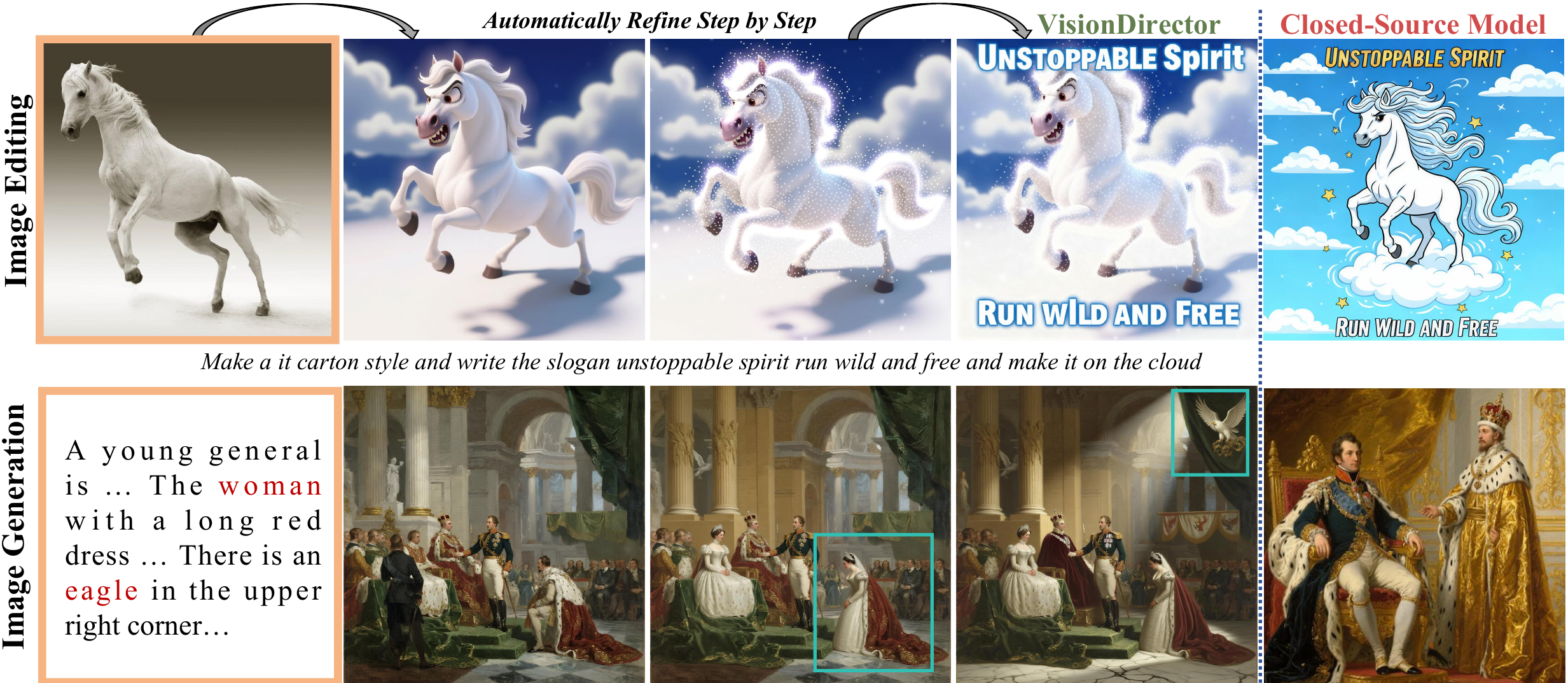}
\caption{VisionDirector is a framework that utilizes the VLM Planner to decompose tasks into multiple goals, perform planning and judgment, and progressively optimize both image-editing and image-generation tasks. It achieves performance comparable to, and in some cases even surpassing, closed-source commercial models.}
\end{center}%
}]

\def\customfootnotetext#1#2{{%
  \let\thefootnote\relax
  \footnotetext[#1]{#2}}}
\customfootnotetext{1}{\textsuperscript{*}Co-corresponding Authors.}
\customfootnotetext{2}{\textsuperscript{\#}Project Lead.}

\input{sec/0_abstract}    
\input{sec/1_intro}
\input{sec/2_bench}

\input{sec/3_methods}
\input{sec/4_experiment}
\input{sec/5_discussions}
\input{sec/6_related}

\input{sec/7_conclusion}
{
    \small
    \bibliographystyle{ieeenat_fullname}
    \bibliography{main}
}


\end{document}

%% file: sec/0_abstract.tex
\begin{abstract}
Generative models can now produce photorealistic imagery, yet they still struggle with the long, multi-goal prompts that professional designers issue. To expose this gap and better evaluate models’ performance in real-world, we introduce Long Goal Bench~(\textbf{LGBench}), a 2{,}000-task suite (1{,}000 T2I, 1{,}000 I2I) whose average instruction contains 18---22 tightly coupled goals spanning global layout, local object placement, typography, and logo fidelity. We find even state-of-the-art models satisfy fewer than 72\% of the goals and routinely miss localized edits, confirming the brittleness of current pipelines. To address this, we present \textbf{VisionDirector}, a training-free, vision-language supervisor that (i) extracts structured goals from long instructions, (ii) dynamically decides between one-shot generation and staged edits, (iii) runs micro-grid sampling plus semantic verification/rollback after every edit, and (iv) logs goal-level rewards. We further fine-tune the planner with Group Relative Policy Optimization, yielding shorter edit trajectories (3.1 vs.\ 4.2 steps) and stronger alignment. VisionDirector achieves new state of the art on GenEval (+7\% overall) and ImgEdit (+0.07 absolute) while producing consistent qualitative improvements on typography, multi-object scenes, and pose editing. 
\vspace{-0.5cm}
\end{abstract}

%% file: sec/1_intro.tex
\section{Introduction}
\label{sec:intro}
The field of visual content creation has advanced rapidly with the rise of diffusion models, which now enable a wide range of applications in image generation and editing. However, professional designers still struggle to achieve faithful results because real-world design briefs often consist of \textbf{long paragraphs with multiple goals}, combining global art direction with numerous local constraints on characters, lighting, typography, and logo placement.
However, existing benchmarks such as DrawBench\cite{saharia2022photorealistictexttoimagediffusionmodels}, TIFA\cite{hu2023tifa}, and MagicBrush\cite{zhang2023magicbrush} predominantly use simplified prompts with at most one or two explicit goals, thereby masking model brittleness in realistic, multi-objective scenarios.

To expose this gap and better evaluate models' performance in real-world production settings, we introduce \textbf{LongGoalBench (LGBench)}, which consists of 2,000 tasks sourced from artist workflows and includes multiple real-world targets per prompt.
Specifically, each text-to-image (T2I) prompt specifies an average of 18.0 goals across 200 categories and 418 subcategories, for example, nature landscapes and historical scene, while the image-to-image (I2I) split, based on real source photos, contains an average of 11.2 edit directives.
We evaluate several state-of-the-art  models, including Flux-Krea \cite{flux1kreadev2025}, Flux-Kontex \cite{flux2024}, and Qwen-Image \cite{wu2025qwen}, on our proposed LGBench. As shown in Table~\ref{tab:model_performance}, even the leading generative models, such as Flux-Kontex and Qwen-Image, only achieves 55.9\% and 71.8\% finish score, respectively. And they fail even more frequently on local text and logo edits. 
These results reveal why current models often fall short in real-world production settings. Although they demonstrate strong aesthetic abilities and can effectively handle single-goal instructions, they struggle with long and multi-goal instructions. Hence, a natural question arises: \textit{how can we enhance their ability to process multi-goal instructions without compromising their aesthetic quality?}

To solve this question, we propose \textbf{VisionDirector}, a training-free, VLM-guided \cite{yang2025visionthink,chu2024towards} controller that decomposes long prompts into structured goals, decides whether to solve them in one shot or via staged micro-edits, and verifies/rolls back every edit. Instead of reporting a single reward at the end of generation, VisionDirector inspects each iteration, maintains goal-level state, and emits short instructions to the underlying diffusion models. This design lets us keep high-quality pretrained editors intact while adding a reasoning layer on top. Nevertheless, the controller can still be refined: we further fine-tune the planner with Group Relative Policy Optimization (GRPO), letting the agent experience thousands of long-goal rollouts and quickly learn when to STOP, VERIFY, or EDIT. By applying our proposed VisionDirector framework in post-training, the model is able to generate the desired images more quickly, reducing the average number of generation rounds from 4.2 to 3.1, achieving about a 26\% reduction.

Our contributions are three-fold:
\begin{itemize}
    \item \textbf{Benchmark.} To bridge the gap between traditional benchmarks and real-world design workflows, we propose LongGoalBench, a benchmark comprising 2{,}000 long-form tasks with 29{,}252 annotated goals, dual T2I/I2I modalities, and automated goal-level verification. We release all prompts, reference images, and evaluation scripts to facilitate reproducible future research.
    \item \textbf{Framework.} VisionDirector provides a modular closed loop consisting of instruction intake, goal planning, one-shot gating, micro-grid execution, VLM verification/rollback, and decision policies. It improves multi-goal adherence without retraining diffusion backbones.
    \item \textbf{RL fine-tuning.} We adapt GRPO to this visual editing context by introducing token-level masking, an alignment-based reward, and rollout workers that render images for every sampled policy. The RL-tuned planner reduces editing rounds by roughly 26\% while boosting GenEval and ImgEdit scores.
\end{itemize}

Taken together, LGBench exposes the challenging cases that current models overlook, VisionDirector addresses these challenges through a director-style agent, and GRPO further refines the policy to produce reliable, instruction-following imagery.

%% file: sec/2_bench.tex
\section{LGBench: LongGoal Benchmark}
\label{sec:lgb}

\subsection{Benchmark Overview}
\label{sec:lgb-overview}

\begin{figure}[t]
  \centering
  \includegraphics[width=0.99\linewidth]{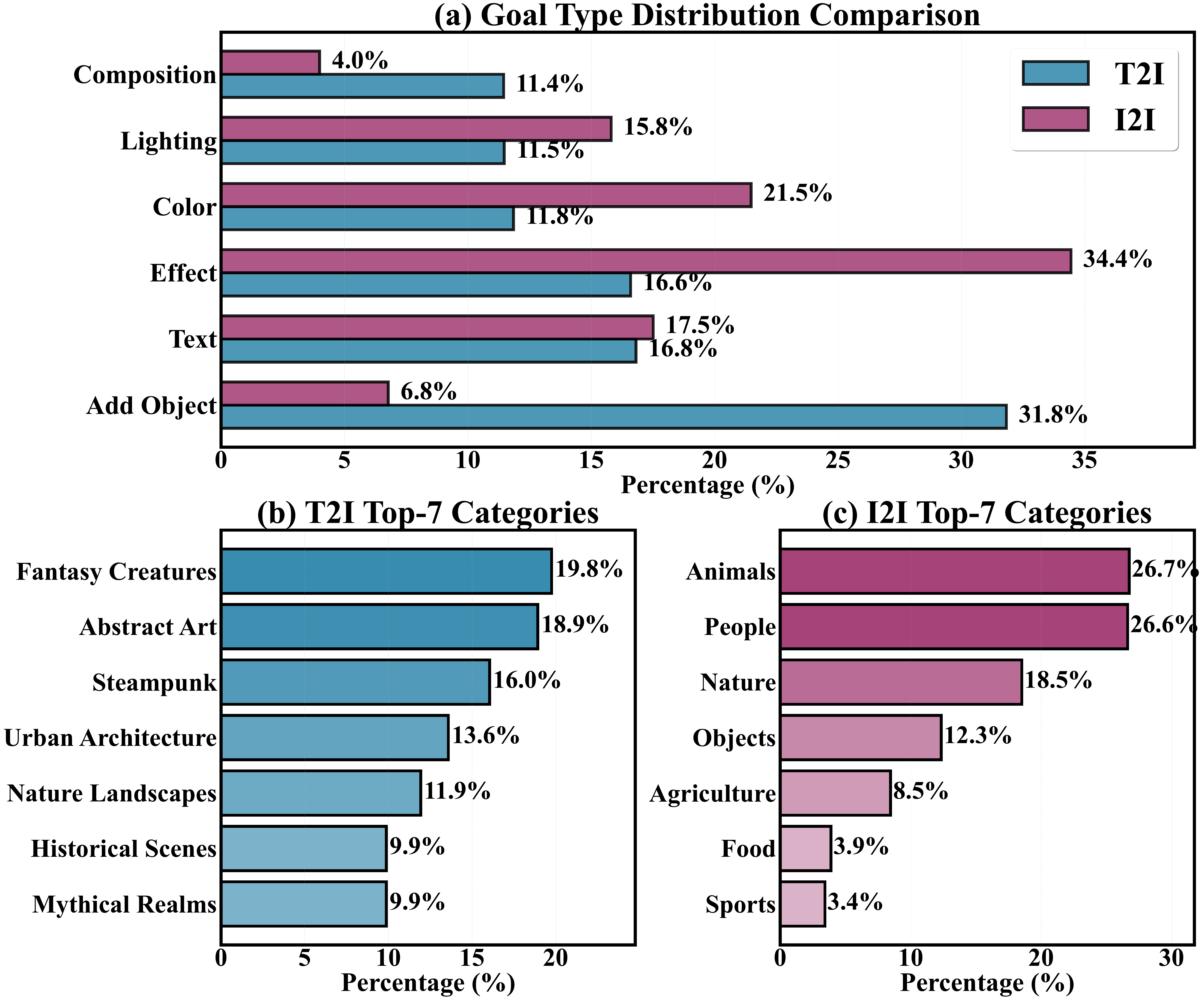}
  \caption{\textbf{Distribution of goal types in LGBench for both T2I and I2I subsets.} The benchmark balances additive, stylistic, and semantic directives, reflecting the multi-constraint nature of real design tasks.}
  \label{fig:lgb_pie}
  \vspace{-8mm}
\end{figure}

To bridge the gap between the complex, multi-goal requirements of real-world design workflows and the limited scope of existing benchmarks, we introduce \textbf{LongGoalBench (LGBench)}.
Unlike prior datasets that evaluate single attributes in isolation, LGBench simulates the actual challenges creative teams face when composing hero shots or refining marketing assets, where success depends on meeting dozens of stylistic, structural, and semantic goals simultaneously.

\paragraph{Scale and coverage.}
As shown in Figure \ref{fig:lgb_pie} and Table \ref{tab:dataset_statistics}, LGBench contains $2000$ tasks in total---$1000$ for text-to-image (T2I) generation and $1000$ for image-to-image (I2I) editing---covering $29{,}252$ individual goal statements.
The T2I subset spans $200$ categories and $418$ fine-grained subcategories (e.g., fantasy\_creatures, urban\_architecture), averaging $18.0$ goals per prompt (range $15$--$23$).
Goal types are balanced across additive objects (31.8\%), textual overlays (16.8\%), visual effects (16.6\%), color constraints (11.8\%), and lighting directives (11.5\%), ensuring diverse coverage of stylistic and compositional aspects.
The I2I subset focuses on realistic photo edits across $29$ coarse classes (densely covering animals, people, and nature), mapped to roughly $710$ subcategories.
Each I2I instruction includes $11.2$ goals on average (range $10$--$22$), with emphasis on effects (34.4\%), color grading (21.5\%), typography (17.5\%), and lighting refinement (15.8\%).



\subsection{Benchmark Construction}
\label{sec:lgb-construction-final}

\begin{figure}[t]
  \centering
  \includegraphics[width=0.99\linewidth]{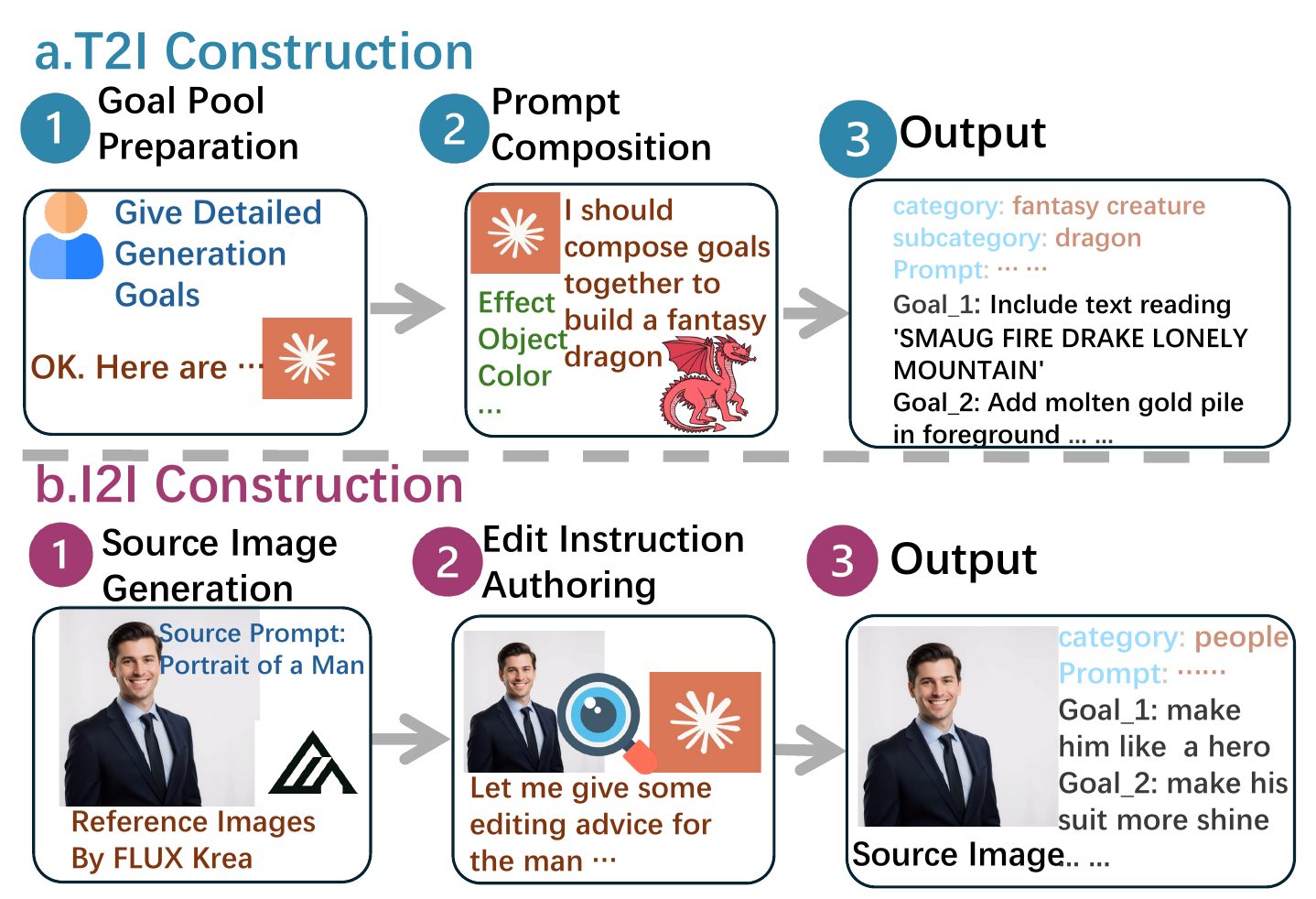}
  \caption{\textbf{Construction pipelines for LongGoalBench.}}
  \label{fig:bench_construction}
  \vspace{-0.5cm}
\end{figure}

To scale dataset creation beyond hand-written prompts, we employ a semi-automated pipeline powered by Claude~4.5~Sonnet through our VisionDirector framework showing in Figure \ref{fig:bench_construction}.
Claude serves as a prompt composer: it reads structured goal inventories, reasons about dependencies, and produces coherent, machine-verifiable instructions in natural language.

\mypara{T2I generation pipeline.}
\textbf{(1) Goal pool preparation.}
For each of the 200 high-level categories (e.g., fantasy creatures, urban architecture), we define a concise inventory of fine-grained visual goals such as ``add molten-gold pile at high prominence'' or ``set color temperature to 3200~K''.
Each goal is annotated with a numeric strength (0--100\%) and a goal type (object, text, color, lighting, or effect).

\noindent\textbf{(2) Prompt composition via Claude.}
Claude receives the goal list together with category metadata, then organizes and merges all goals into a single coherent instruction that includes quantitative cues (e.g., ``at moderate intensity (45\%)'') while avoiding contradictions.
Each resulting prompt record contains the category labels, the composed long-form instruction, and structured annotations for individual goals.

\noindent\textbf{(3) Statistics.}
The final 1{,}000 T2I prompts span 418 subcategories and include $18 \pm 3$ goals per prompt, contributing approximately 18k individual objectives.

\mypara{I2I editing pipeline.}
\textbf{(1) Seed image generation.}
We define 29 coarse scene classes covering people, animals, and nature, and generate a representative base image for each using the Flux Krea model as the editable ``before'' reference.

\noindent\textbf{(2) Edit instruction authoring.}
Claude (vision-enabled) is provided with the base image and its class description.
It analyzes salient regions---for instance, detecting faces, lighting, or text overlays---and formulates 10--22 concrete edit directives such as ``increase hair shine by 60\%'' or ``overlay TIMELESS BEAUTY in elegant serif font''.
Because Claude observes the actual image, the edits it proposes remain spatially and semantically consistent with the layout.

\noindent\textbf{(3) Data structure.}
Each I2I instruction entry stores the class metadata, long-form directive, and structured goal descriptors, while the paired source images are linked via unique identifiers.
The I2I subset contributes 11,217 additional goals, yielding a total of 29{,}252 annotated objectives across both modalities.

\begin{table}[t]
\centering
\footnotesize

\begin{tabular*}{\linewidth}{@{\extracolsep{\fill}} lcc}
\toprule
\textbf{Metric} & \textbf{T2I} & \textbf{I2I} \\
\midrule
\multicolumn{3}{l}{\textit{Overall Statistics}} \\
Total Samples & 1,000 & 1,000 \\
Total Categories & 200 & 29 \\
Total Subcategories & 418 & 710 \\
Avg. Prompt Length & 207.6 & 253.4 \\
Avg. Goals/Sample & 18.04 & 11.22 \\
Total Goals & 18,035 & 11,217 \\
\midrule
\multicolumn{3}{l}{\textit{Goal Type Count}} \\
Add Object & 5,739 & 760 \\
Text & 3,032 & 1,964 \\
Effect & 2,993 & 3,863 \\
Color & 2,137 & 2,409 \\
Lighting & 2,069 & 1,773 \\
Composition & 2,065 & 448 \\
\bottomrule
\end{tabular*}
\caption{\textbf{Statistics result of the LGBench.}}
\label{tab:dataset_statistics}
\vspace{-0.5cm}
\end{table}

\begin{table*}[t]
  \centering
  \small
  \setlength{\tabcolsep}{5pt}

  \begin{tabular*}{\textwidth}{@{\extracolsep{\fill}} lcccccccccc}
    \toprule
    Model & Modal & Finish & Success\textsubscript{$\geq 80$\%} & AddObj & Text & Effect & Color & Light & Composition\\
    \midrule
    \multicolumn{10}{l}{\textit{Base models (no VisionDirector)}} \\
    \midrule
    Qwen-Image (T2I) ~\cite{wu2025qwen}        & T2I & 71.8 & 30.7 & 76.6 & 29.7 & 87.0 & 78.3 & 85.8 & 77.5\\
    Flux-Krea (T2I) ~\cite{flux1kreadev2025}   & T2I & 66.8 & 18.6 & 75.7 & 25.6 & 82.3 & 67.7 & 78.1 & 67.8\\
    Flux-Dev (T2I)  ~\cite{flux2024}           & T2I & 40.0 &  1.0 & 32.2 &  0.8 & 55.4 & 58.2 & 62.9 & 55.1\\
    Qwen-Edit+ (I2I) ~\cite{wu2025qwen}        & I2I & 71.0 & 52.4 & 62.0 & 93.2 & 60.3 & 77.0 & 70.9 & 49.3\\
    Qwen-Edit (I2I)  ~\cite{wu2025qwen}        & I2I & 65.3 & 40.1 & 56.7 & 66.4 & 60.1 & 74.6 & 69.3 & 54.1\\
    Flux-Kontext (I2I) ~\cite{batifol2025flux} & I2I & 55.9 & 31.0 & 49.7 & 70.8 & 48.3 & 62.6 & 53.6 & 39.7\\
    \midrule
    \multicolumn{10}{l}{\textit{Base + VisionDirector (VDred: absolute gain over base, per goal type)}} \\
    \midrule
    Qwen-Image + VID   & T2I
      & 74.8\textsubscript{\textcolor{VDred}{+3.0}}
      & 36.7\textsubscript{\textcolor{VDred}{+6.0}}
      & 79.1\textsubscript{\textcolor{VDred}{+2.5}}
      & 37.2\textsubscript{\textcolor{VDred}{+7.5}}
      & 88.4\textsubscript{\textcolor{VDred}{+1.4}}
      & 80.6\textsubscript{\textcolor{VDred}{+2.3}}
      & 87.3\textsubscript{\textcolor{VDred}{+1.5}}
      & 79.9\textsubscript{\textcolor{VDred}{+2.4}}\\

    Flux-Krea + VID    & T2I
      & 74.9\textsubscript{\textcolor{VDred}{+8.1}}
      & 35.9\textsubscript{\textcolor{VDred}{+17.3}}
      & 81.6\textsubscript{\textcolor{VDred}{+5.9}}
      & 43.7\textsubscript{\textcolor{VDred}{+18.1}}
      & 86.6\textsubscript{\textcolor{VDred}{+4.3}}
      & 75.6\textsubscript{\textcolor{VDred}{+7.9}}
      & 83.4\textsubscript{\textcolor{VDred}{+5.3}}
      & 75.7\textsubscript{\textcolor{VDred}{+7.9}}\\

    Flux-Dev + VID     & T2I
      & 62.4\textsubscript{\textcolor{VDred}{+22.4}}
      & 25.6\textsubscript{\textcolor{VDred}{+24.6}}
      & 57.5\textsubscript{\textcolor{VDred}{+25.3}}
      & 29.5\textsubscript{\textcolor{VDred}{+28.7}}
      & 75.9\textsubscript{\textcolor{VDred}{+20.5}}
      & 77.6\textsubscript{\textcolor{VDred}{+19.4}}
      & 80.1\textsubscript{\textcolor{VDred}{+17.2}}
      & 74.0\textsubscript{\textcolor{VDred}{+18.9}}\\

    Qwen-Edit+ + VID   & I2I
      & 76.7\textsubscript{\textcolor{VDred}{+5.7}}
      & 63.5\textsubscript{\textcolor{VDred}{+11.1}}
      & 69.5\textsubscript{\textcolor{VDred}{+7.5}}
      & 94.5\textsubscript{\textcolor{VDred}{+1.3}}
      & 68.1\textsubscript{\textcolor{VDred}{+7.8}}
      & 81.5\textsubscript{\textcolor{VDred}{+4.5}}
      & 76.6\textsubscript{\textcolor{VDred}{+5.7}}
      & 59.3\textsubscript{\textcolor{VDred}{+10.0}}\\

    Qwen-Edit + VID    & I2I
      & 70.6\textsubscript{\textcolor{VDred}{+5.3}}
      & 56.4\textsubscript{\textcolor{VDred}{+16.3}}
      & 63.3\textsubscript{\textcolor{VDred}{+6.6}}
      & 71.5\textsubscript{\textcolor{VDred}{+5.1}}
      & 66.2\textsubscript{\textcolor{VDred}{+6.1}}
      & 78.5\textsubscript{\textcolor{VDred}{+3.9}}
      & 74.0\textsubscript{\textcolor{VDred}{+4.7}}
      & 61.1\textsubscript{\textcolor{VDred}{+7.0}}\\

    Flux-Kontext + VID & I2I
      & 62.4\textsubscript{\textcolor{VDred}{+6.5}}
      & 49.7\textsubscript{\textcolor{VDred}{+18.7}}
      & 57.1\textsubscript{\textcolor{VDred}{+7.4}}
      & 75.1\textsubscript{\textcolor{VDred}{+4.3}}
      & 55.9\textsubscript{\textcolor{VDred}{+7.6}}
      & 68.1\textsubscript{\textcolor{VDred}{+5.5}}
      & 60.4\textsubscript{\textcolor{VDred}{+6.8}}
      & 48.6\textsubscript{\textcolor{VDred}{+8.9}}\\
    \bottomrule
  \end{tabular*}
  \vspace{-0.2cm}
  \caption{\textbf{Performance comparison of models with and without VisionDirector.}}
  \label{tab:model_performance}
\end{table*}
\begin{table*}[t]
  \centering
  \small
  \setlength{\tabcolsep}{5pt}
  \begin{tabular*}{0.97\textwidth}{@{\extracolsep{\fill}} lc|ccccc}
    \toprule
    Property & LGBench & DrawBench~\cite{saharia2022photorealistic} & TIFA~\cite{hu2023tifa} & GenEval~\cite{ghosh2023geneval} & MagicBrush~\cite{zhang2023magicbrush} & ImgEdit~\cite{ye2025imgedit} \\
    \midrule
    Dual Modalities (T2I + I2I)          & \cmark & \xmark & \xmark & \xmark & \xmark & \xmark \\
    Long-chain Instructions (10$+$ goals) & \cmark & \xmark & \xmark & \xmark & \xmark & \xmark \\
    Hybrid-goal Tasks ($>5$ different goals)        & \cmark & \xmark & \xmark & \xmark & \xmark & \xmark \\
    Automated Goal-level Verification    & \cmark & \xmark & \xmark & \xmark & \xmark & \xmark \\

    \bottomrule
  \end{tabular*}
 \vspace{-0.2cm}
  \caption{\textbf{Comparison between LGBench and other T2I/I2I benchmarks.}}
  \label{tab:lgb-compare}
  \vspace{-0.5cm}
\end{table*}

\mypara{Automated execution.}
All tasks are generated and evaluated automatically through a distributed execution pipeline that assigns workloads across GPUs, loads the designated diffusion or editing backbone (e.g., Flux, Qwen), and enforces consistent resolution and formatting.
Each run outputs unified image, metadata, and prompt files in a standardized format.
This modular design allows new generative models to be integrated seamlessly without re-annotation or manual intervention.

\subsection{Evaluation Results}
\label{sec:lgb-eval-final}

Each generated image or edited result is automatically evaluated by a multimodal verifier built on Qwen3-VL-32B-Instruct.
For image-to-image tasks, the verifier compares the ``before'' and ``after'' images; for text-to-image tasks, it inspects the generated output directly.
Long-form prompts are internally converted into structured instructions so that the verifier can reason step by step, produce explanations, and return verdicts for each goal.
Responses with confidence below 0.81 are filtered out.
A task is labeled as success if at least 80\% of its goals are satisfied, partial if coverage lies between 0--80\%, and failure if no goal passes.
We also compute per-goal-type pass rates to diagnose which visual dimensions degrade first.

Across both tracks, Qwen-based generators exhibit the strongest multi-goal consistency.
In the text-to-image track, \textbf{Qwen-Image} achieves a 71.8\% overall goal success rate, clearing the 80\% threshold on nearly one third of all prompts.
\textbf{Flux-Krea} trails by roughly five percentage points, while \textbf{Flux-Dev} collapses to 40.0\% and almost never satisfies text-related goals (0.8\% success).
On the editing side, \textbf{Qwen-Edit+} attains 71.0\% goal success with more than half of its tasks fully completed, outperforming both \textbf{Qwen-Edit} (65.3\%) and \textbf{Flux-Kontext} (55.9\%), the latter showing a 7.3\% failure rate.

These trends reveal consistent bottlenecks in \emph{typography}, \emph{additive object placement}, and \emph{lighting refinement}—dimensions often ignored by single-score benchmarks.
We plan to extend evaluations to additional state-of-the-art and closed-source systems (e.g., HiDream, Janus, Bagel, Nana Banana, and SeedDream~4.0) once access becomes available.
Overall, the results confirm that LGBench effectively exposes where current diffusion and editing models break down under long, interdependent goals, motivating the need for a higher-level \emph{director} that can reason about multi-step objectives.
This insight directly inspires the design of \textbf{VisionDirector}, introduced in the next section.
\begin{figure*}[!t]
  \centering
  \includegraphics[width=0.87\linewidth]
{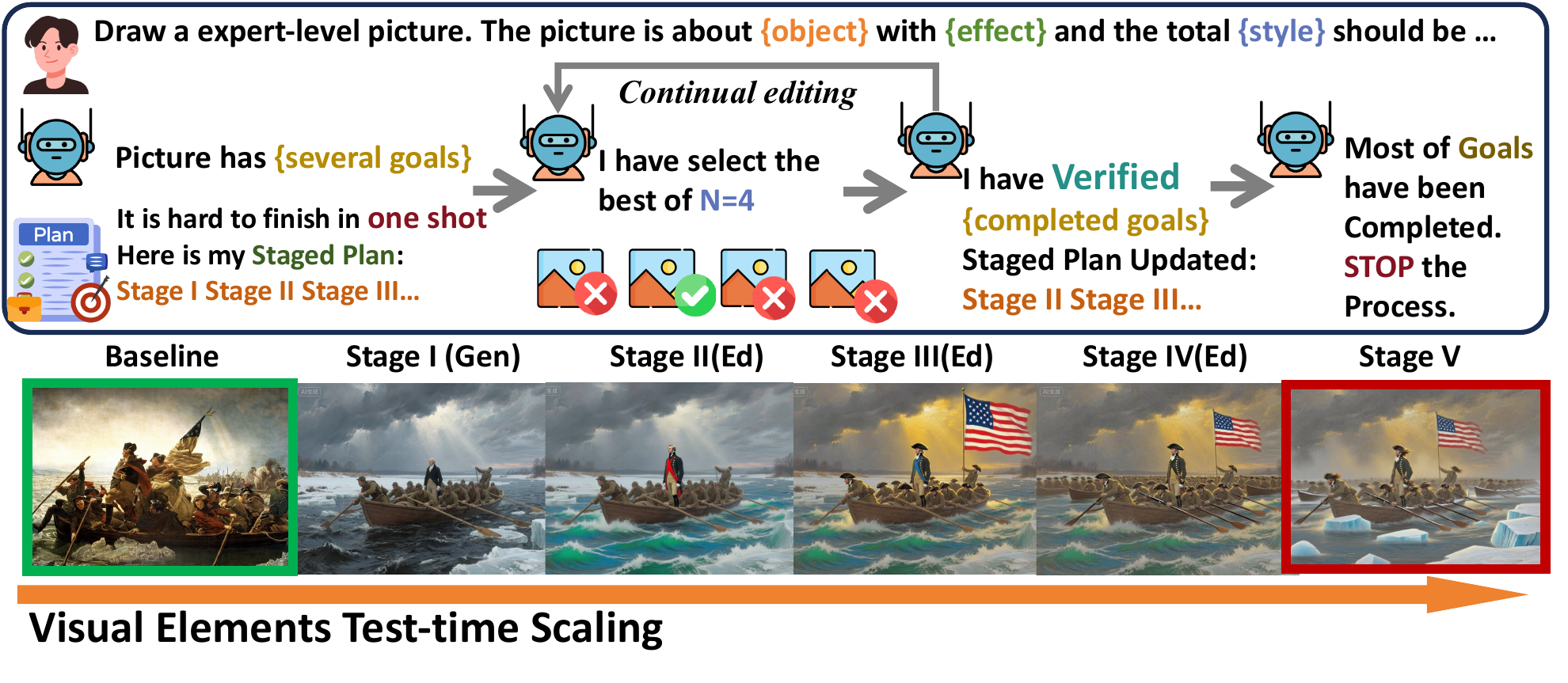}
  \caption{\textbf{Workflow of VisionDirector.} The planner interprets long instructions, decides between one-shot or staged execution, performs micro-grid sampling, verifies progress, and rolls back if needed.}
  \label{fig:director}
  \vspace{-0.5cm}
\end{figure*}

\subsection{Discussion and Motivation}
LGBench is designed to evaluate \emph{multi-attribute alignment} rather than simple prompt fidelity.
In real-world creative production, tasks rarely involve the one-shot fulfillment of a single textual cue; instead, they require the coherent realization of multiple interdependent goals such as composition, lighting, color harmony, and textual consistency. However, existing evaluation suites offer limited insight into this regime. 
As summarized in Table~\ref{tab:lgb-compare}, previous benchmarks such as DrawBench~\cite{saharia2022photorealistic}, and TIFA~\cite{hu2023tifa} rely on short prompts and human preference studies. GenEval~\cite{ghosh2023geneval} and MagicBrush~\cite{zhang2023magicbrush} introduce a limited number of QA-style or short edit tasks but still emphasize subjective or low-granularity scoring. None provide large-scale, reproducible evaluations in which success depends on meeting 10–23 quantitative goals verified automatically.

LGBench fills this gap by expanding both the instruction space and the scoring granularity. Each task consists of long, structured goal chains authored by Claude, paired with Flux-generated reference images and verified through a Qwen-based vision–language pipeline.
This design enables the systematic measurement of fine-grained attributes—for example, assessing whether typography is legible and properly positioned, or whether the light–color balance and overall composition remain consistent after multiple edits.
By quantifying such aspects, LGBench reveals nuanced failure modes in modern diffusion and editing systems that were previously anecdotal, such as neglecting embedded text or over-saturating local illumination.

Beyond ranking models, LGBench serves as infrastructure for studying \emph{goal-conditioned training}, \emph{adaptive guidance}, and \emph{verification-aware generation}. 
Practitioners can examine per-goal statistics to identify where models fail, then refine training data, negative prompts, or guidance schedules accordingly.
As shown in Table.~\ref{tab:model_performance}, even state-of-the-art commercial systems repeatedly miss local text or logo constraints when instructions contain 15 or more interacting goals, underscoring the need for a director-style agent to manage multi-goal consistency.
Hence, we try to propose a solution to solve this challenge in Sec.~\ref{sec:visiondirector}.

\begingroup
\setlength{\tabcolsep}{3.5pt}         
\renewcommand{\arraystretch}{0.95}     
\captionsetup{skip=2pt}                

\begin{table}[t]
  \centering
  \footnotesize
  \resizebox{\linewidth}{!}{%
  \begin{tabular}{lccccccc}
    \toprule
    Model & Single Obj. & Two Obj. & Counting & Colors & Position & Attribute & Overall$\uparrow$ \\
    \midrule
    Show-o ~\cite{xie2024show} & 0.95 & 0.52 & 0.49 & 0.82 & 0.11 & 0.28 & 0.53 \\
    Emu3-Gen~\cite{wang2024emu3} & 0.98 & 0.71 & 0.34 & 0.81 & 0.17 & 0.21 & 0.54 \\
    PixArt-$\alpha$ ~\cite{chen2023pixart}& 0.98 & 0.50 & 0.44 & 0.80 & 0.08 & 0.07 & 0.48 \\
    SD3 Medium~\cite{esser2024scaling} & 0.98 & 0.74 & 0.63 & 0.67 & 0.34 & 0.36 & 0.62 \\
    FLUX.1 [Dev] ~\cite{flux2024}& 0.98 & 0.81 & 0.74 & 0.79 & 0.22 & 0.45 & 0.66 \\
    SD3.5 Large ~\cite{esser2024scalingrectifiedflowtransformers}& 0.98 & 0.89 & 0.73 & 0.83 & 0.34 & 0.47 & 0.71 \\
    JanusFlow ~\cite{ma2025janusflow}& 0.97 & 0.59 & 0.45 & 0.83 & 0.53 & 0.42 & 0.63 \\
    Lumina-Image 2.0 ~\cite{qin2025lumina} & 1.00 & 0.87 & 0.67 & 0.93 & 0.34 & 0.62 & 0.73 \\
    Janus-Pro-7B ~\cite{chen2025janus}& 0.99 & 0.89 & 0.59 & 0.90 & 0.79 & 0.66 & 0.80 \\
    HiDream-I1-Full ~\cite{cai2025hidream}& \textbf{1.00} & 0.98 & 0.79 & 0.91 & 0.60 & 0.72 & 0.83 \\
    GPT Image 1 [High] ~\cite{openai2024gpt4technicalreport}& 0.99 & 0.92 & 0.85 & 0.92 & 0.75 & 0.61 & 0.84 \\
    Seedream 3.0 ~\cite{gao2025seedream} & 0.99 & \textbf{0.96} & 0.91 & \textbf{0.93} & 0.47 & 0.80 & 0.84 \\
    Qwen-Image ~\cite{wu2025qwen}& 0.99 & 0.92 & 0.89 & 0.88 & 0.76 & 0.77 & 0.87 \\
    \midrule
    VisionDirector & 0.99 & 0.94 & \textbf{0.96} & 0.92 & \textbf{0.88} & \textbf{0.95} & \textbf{0.94} \\
    \bottomrule
  \end{tabular}}

  \caption{\textbf{Results on GenEval Bench.} ``Single Obj.'' and ``Two Obj.'' denote single- and two-object prompts. The ``Overall'' column averages the scores across the nine tasks.}
  \label{tab:geneval}
  \vspace{-0.5cm}
\end{table}
\endgroup

\begin{figure*}[t]
      \centering
      \includegraphics[width=0.87\linewidth]{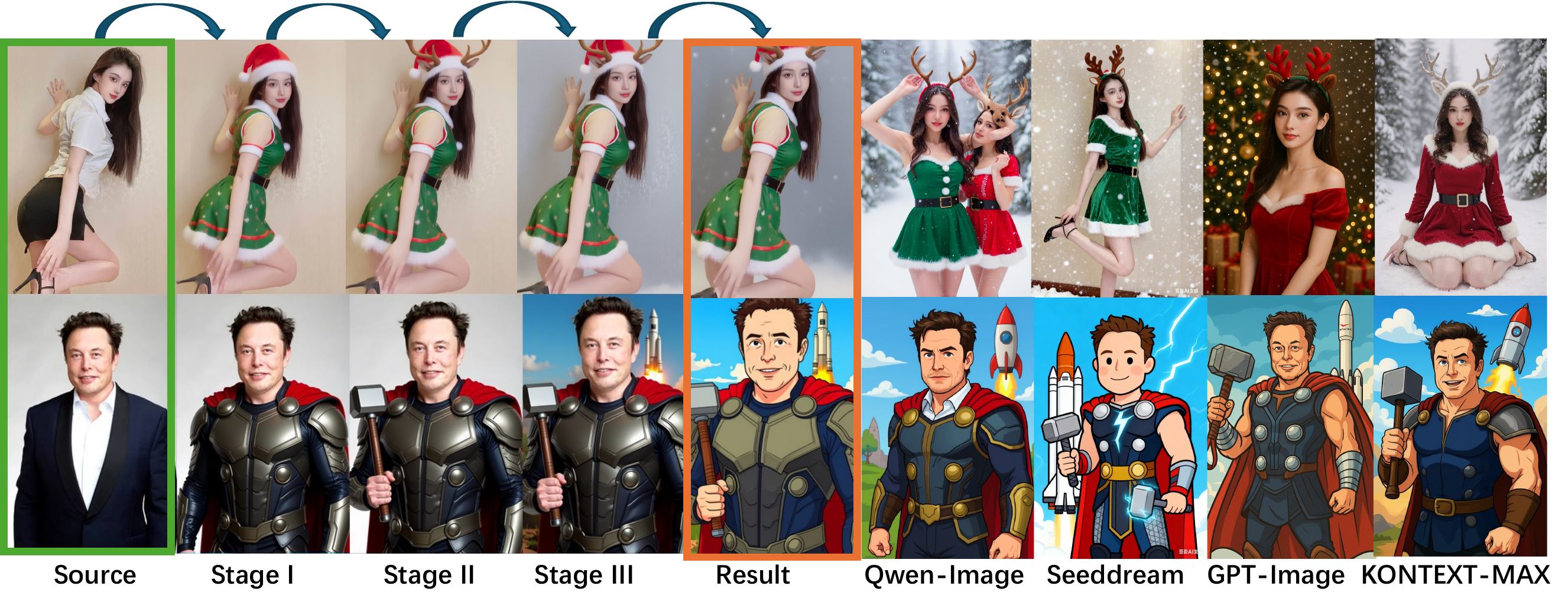}
      \caption{\textbf{I2I result comparison with other models.}}
    \label{fig:i2i_qual}
    \vspace{-0.5cm}
  \end{figure*}

%% file: sec/3_methods.tex
\section{VisionDirector}
\label{sec:visiondirector}

\subsection{Overview}

Real-world creative tasks often require iterative coordination rather than one-shot generation.
\textbf{VisionDirector} instantiates a closed-loop ``director'' that plans, executes, and verifies visual edits under long, interdependent instructions.
The system integrates three modular components:
(1) a \emph{planner VLM} (Qwen3-VL-8B \cite{bai2025qwen3vltechnicalreport} ) that parses and schedules goals,
(2) \emph{text-to-image} and \emph{image-to-image} editors (Qwen-Image, Qwen-Image-Edit) that perform visual updates, and
(3) a \emph{verifier VLM} that provides goal-level feedback.
The planner owns the semantics: it decomposes the user instruction into structured goals, decides execution order, and supervises revisions until convergence.
Editors are stateless executors that follow the planner’s directives, while the verifier evaluates results and signals when to stop.
Because each module communicates through natural-language messages, components can be swapped or upgraded without retraining the planner, and the same control policy can later be fine-tuned through reinforcement learning (§\ref{sec:grpo}).

\subsection{Training-Free Closed-Loop Control}

Figure~\ref{fig:director} illustrates the overall workflow.
The training-free controller operates through eight deterministic stages inspired by Deepseek-R1-style \cite{guo2025deepseek} reasoning but specialized for visual generation.


\mypara{\textit{Step1: Instruction intake.}}
A raw multimodal instruction (optionally with a reference image) enters the system.
Basic metadata such as resolution and language tags are normalized before reasoning.

\mypara{\textit{Step2: Goal extraction and planning.}}
The planner extracts goals \emph{verbatim} from the instruction, assigning each a type (\emph{global}, \emph{local}, \emph{text}, \emph{layout}, etc.), a conflict flag, and a scalar score estimating one-shot feasibility.
All goals are stored in a pending list; a separate completed set starts empty.

\mypara{\textit{Step3: One-shot or staged execution.}}
If the estimated one-shot score is high and the affected area small, the planner directly attempts a complete composition.
Otherwise, it schedules goals in batches of one or two, ordered from global to local constraints, so that each edit remains focused.

\mypara{\textit{Step4: Micro-grid execution.}}
For each active batch, the planner issues a concise directive (e.g., ``Raise the key light on the hero’s left cheek'').
Multiple candidates are generated with different random seeds, and a lightweight VLM judge selects the best image.
This mitigates diffusion randomness without significant latency.

\mypara{\textit{Step5: Verification and rollback.}}
A verifier VLM evaluates results against pending goals, marking each satisfied or not. If the new edit degrades overall alignment, the system reverts to the previous best image and reshuffles remaining goals to avoid repeating the same conflict.

\mypara{\textit{Step6: Decision and continuation.}}
If all goals are satisfied, the loop terminates.
Otherwise, the planner adapts the editing mode: local adjustments trigger inpainting; unsatisfied global constraints prompt scene-level regeneration.
Every few rounds, a self-query (“Can the image still improve?”) triggers early stopping if confidence is low.

\mypara{\textit{Step7\&8: Looping and outputs.}}
The loop continues until no further improvements are detected.
The final image and full reasoning trace (goals, edits, verdicts) are returned for transparent evaluation and auditing.

\subsection{GRPO for Efficient Generation}
\label{sec:grpo}

While the training-free VisionDirector already enhances instruction following, it still requires multiple rounds of editing before convergence. Furthermore, VisionDirector is not only a training-free pipeline but can also be employed after training to further improve generation efficiency.
Specifically, we post-train the planner VLM using \textbf{Group Relative Policy Optimization (GRPO)}~\cite{shao2024deepseekmath}, a reinforcement learning method that encourages concise, high-reward edit strategies.

Let \(x\) denote the multimodal prompt and \(y=(y_1,\ldots,y_T)\) the planner’s interleaved \emph{Describe–Inspect–Revise} actions.
We train the planner with a GRPO variant of PPO over \(G\) sampled trajectories per prompt:

\begingroup
\setlength{\abovedisplayskip}{4pt}
\setlength{\belowdisplayskip}{4pt}
\footnotesize
\begin{equation}
\label{eq:grpo-vision}
\begin{aligned}
\mathcal{J}_{\text{GRPO}}(\theta)
= \mathbb{E}_{x,\{y^{(i)}\}}\Bigg[
&\frac{1}{G}\sum_{i=1}^{G}
  \frac{1}{\sum_{t} I(y^{(i)}_{t})}
  \sum_{t} I(y^{(i)}_{t})\;
  \mathcal{L}_{\text{clip}}\!\big(\rho^{(i)}_{t},\,\hat{A}^{(i)}_{t}\big)
\\
&\quad-\;
\beta\,\mathrm{KL}\!\big(\pi_{\theta}\,\|\,\pi_{\text{ref}}\big)
\Bigg],
\end{aligned}
\end{equation}
\endgroup
where \(I(\cdot)\) masks tokens originating from external tools so that updates apply only to the planner’s textual actions, and \(\beta\) controls KL regularization toward a reference policy \(\pi_{\text{ref}}\).
Here \(\mathcal{L}_{\text{clip}}\) is the standard PPO-style clipped surrogate loss,
\(\rho^{(i)}_{t}\) is the per-token importance ratio, and \(\hat{A}^{(i)}_{t}\) is the group-normalized advantage using the average group reward as a baseline.

\mypara{Reward model and rollout.}
Each trajectory replays the entire director loop and generates a final edited image.
A separate alignment VLM scores how well the image satisfies the target instruction on a 0--5 scale, providing dense pixel-level supervision for attributes such as typography, lighting, and spatial layout.
Rewards are normalized within each group before policy updates.

\mypara{Effect on inference.}
The GRPO-tuned planner cuts the median editing rounds from 4.2 to 3.1, reducing inference latency by about 26\% without lowering verification accuracy.
Empirically, the optimized policy proposes higher-gain edits earlier and terminates confidently once only minor residual goals remain, effectively complementing the training-free director pipeline above.

%% file: sec/4_experiment.tex
\section{Experiments}
\label{sec:exp}


\subsection{Evaluation Setup}
\mypara{Benchmarks.}
We evaluate VisionDirector on two benchmarks that stress long-horizon instruction following. For text-to-image (T2I), we report accuracy on GenEval~\cite{ghosh2023geneval}. GenEval focuses on compositional prompts with multiple objects and attributes. For image-to-image (I2I) editing we adopt ImgEdit~\cite{ye2025imgedit}, which covers nine edit primitives.

\mypara{Implementation Details}
Unless otherwise stated we run the VisionDirector loop with four T2I samples or a single I2I edit per iteration and cap the closed-loop refinement at six iterations. All competing models use their publicly released settings; tables retain the original leaderboard numbers so that improvements attributable to VisionDirector can be read in-context. Appendix material provides additional ablations on iteration count and verifier confidence thresholds.

\subsection{Results on Text-to-Image}
Table~\ref{tab:geneval} summarizes compositional fidelity on GenEval. VisionDirector, instantiated with a Qwen-Image backbone and the GRPO-trained planning agent, delivers the highest overall score (0.94) while jointly improving all sub-metrics, especially attribute binding and reasoning about relative positions.

\subsection{Results on Image-to-Image}
For editing tasks we follow ImgEdit. ImgEdit scores nine edit primitives from 1--5 via GPT-4.1 judges. VisionDirector surpasses prior open agents on every ImgEdit primitive (Table~\ref{tab:imgedit}) indicating that the closed-loop controller faithfully preserves unedited regions while applying instruction-specific changes.




\begin{table*}[t]
  \centering
  \resizebox{\textwidth}{!}{
  \begin{tabular}{lcccccccccc}
    \toprule
    Model & Add & Adjust & Extract & Replace & Remove & Background & Style & Hybrid & Action & Overall$\uparrow$ \\
    \midrule
    MagicBrush ~\cite{zhang2023magicbrush} & 2.84 & 1.58 & 1.51 & 1.97 & 1.58 & 1.75 & 2.38 & 1.62 & 1.22 & 1.90 \\
    Instruct-Pix2Pix ~\cite{brooks2023instructpix2pix}& 2.45 & 1.83 & 1.44 & 2.01 & 1.50 & 1.44 & 3.55 & 1.20 & 1.46 & 1.88 \\
    AnyEdit ~\cite{jiang2025anyedit} & 3.18 & 2.95 & 1.88 & 2.47 & 2.23 & 2.24 & 2.85 & 1.56 & 2.65 & 2.45 \\
    UltraEdit ~\cite{zhao2024ultraedit}& 3.44 & 2.81 & 2.13 & 2.96 & 1.45 & 2.83 & 3.76 & 1.91 & 2.98 & 2.70 \\
    OmniGen ~\cite{xiao2025omnigen}& 3.47 & 3.04 & 1.71 & 2.94 & 2.43 & 3.21 & 4.19 & 2.24 & 3.38 & 2.96 \\
    ICEdit ~\cite{zhang2025context}& 3.58 & 3.39 & 1.73 & 3.15 & 2.93 & 3.08 & 3.84 & 2.04 & 3.68 & 3.05 \\
    Step1X-Edit ~\cite{liu2025step1x}& 3.88 & 3.14 & 1.76 & 3.40 & 2.41 & 3.16 & 4.63 & 2.64 & 2.52 & 3.06 \\
    BAGEL ~\cite{deng2505emerging}& 3.56 & 3.31 & 1.70 & 3.30 & 2.62 & 3.24 & 4.49 & 2.38 & 4.17 & 3.20 \\
    UniWorld-V1 ~\cite{lin2025uniworld}& 3.82 & 3.64 & 2.27 & 3.47 & 3.24 & 2.99 & 4.21 & 2.96 & 2.74 & 3.26 \\
    OmniGen2 ~\cite{wu2025omnigen2}& 3.57 & 3.06 & 1.77 & 3.74 & 3.20 & 3.57 & 4.81 & 2.52 & 4.68 & 3.44 \\
    FLUX.1 Kontext [Pro] ~\cite{batifol2025flux}& 4.25 & 4.15 & 2.35 & \underline{4.56} & 3.57 & 4.26 & 4.57 & 3.68 & 4.63 & 4.00 \\
    GPT Image 1 [High] ~\cite{openai2024gpt4technicalreport}& \textbf{4.61} & \textbf{4.33} & 2.90 & 4.35 & 3.66 & \textbf{4.57} & \textbf{4.93} & \underline{3.96} & \textbf{4.89} & 4.20 \\
 
    Qwen-Image-Edit ~\cite{wu2025qwen}& 4.38 & 4.16 & \underline{3.43} & \underline{4.66} & \underline{4.14} & 4.38 & 4.81 & 3.82 & 4.69 & \underline{4.27} \\
         \midrule
VisionDirector & \underline{4.55} & \underline{4.32} & \textbf{3.72} & \textbf{4.83} & \textbf{4.41} & \underline{4.52} & \underline{4.92} & \textbf{4.05} & 4.72 & \textbf{4.35} \\
    \bottomrule
  \end{tabular}}
\caption{\textbf{Results on ImgEdit Bench.} The scores range from 1 to 5, representing quality from low to high. The ``Overall'' column averages the scores across the nine tasks. Our VisionDirector significantly outperforms the open-source model and demonstrates competitive performance with the closed-source model.}  \label{tab:imgedit}
\end{table*}

\subsection{Ablations and Analysis}
We ablate core design choices on a held-out LGBench validation split of 200 mixed T2I/I2I tasks.
Table~\ref{tab:grpo_ablate} shows that GRPO shortens trajectories by 26\% while improving goal coverage and reducing edit cost. Removing the micro-grid sampler increases stochastic failures on typography instructions by 7\%, while disabling semantic rollback leads to accumulated hallucinations and a drop in goal coverage to 0.74 (see supplementary for details).

Figure~\ref{fig:vlm_decision} analyzes VLM decision-making under increasing task complexity. The system operates in a high-efficiency regime for tasks with $\leq$15 goals, requiring only 1–3 iterations. Beyond this point, the preference for one-shot execution sharply declines—from over 85\% to below 10\% at 30 goals—indicating an adaptive shift toward staged execution for complex multi-goal tasks.

Table~\ref{tab:ablation} further shows that the refinement strategy outperforms alternative methods, and combining strategies yields the best overall performance.

\begin{figure}[t]
      \centering
      \includegraphics[width=\linewidth]{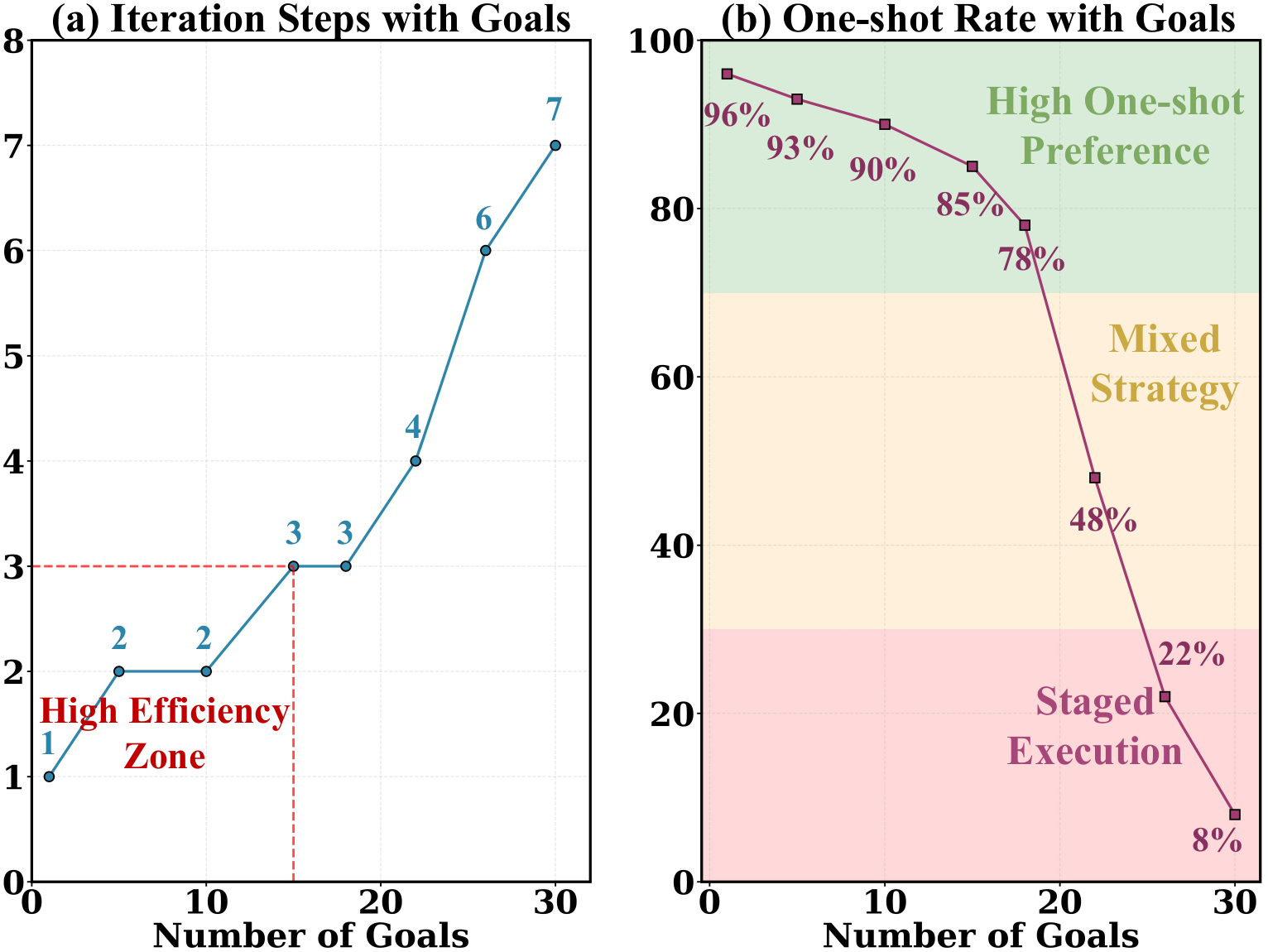}
      \caption{\textbf{Adaptive decision-making in VisionDirector.} The VLM exhibits distinct
   behavioral phases based on task complexity. (a) Iteration steps increase from 1-3 to 5-7
  after exceeding 15 goals. (b) One-shot execution preference ($>$85\%) transitions to staged
   execution ($<$10\%) beyond the critical threshold, demonstrating the system's adaptive
  control strategy.}
      \label{fig:vlm_decision}
      \vspace{-0.5cm}
  \end{figure}

\begin{table}[t]
\small
\setlength{\tabcolsep}{4pt} 
\renewcommand{\arraystretch}{0.9} 
  \centering
  \begin{tabular}{lccc}
    \toprule
    Planner & Steps$\downarrow$ & Goal cov.$\uparrow$ & Edits / task$\downarrow$ \\
    \midrule
    VisionDirector (no RL) & 4.2 & 0.74 & 3.3 \\
    VisionDirector + GRPO  & \textbf{3.1} & \textbf{0.78} & \textbf{2.5} \\
    \bottomrule
  \end{tabular}
    \vspace{-0.2cm}
  \caption{GRPO improves planning efficiency on LGBench. ``Goal cov.'' measures the fraction of verified goals per task; ``Edits'' counts diffusion executions.}
  \label{tab:grpo_ablate}
  \vspace{-0.2cm}
\end{table}

\subsection{Case Study}
Qualitative examples for both modalities are highlighted in Figure.~\ref{fig:i2i_qual}; higher-resolution versions plus interactive galleries will appear on the project page, and additional breakdowns are deferred to the supplementary PDF.

%% file: sec/5_discussions.tex
\section{Discussion}
\subsection{Why we need agent framework for generation?}
The results on LGBench demonstrate that the shortcomings of current leading diffusion models do not stem from a lack of texture fidelity or aesthetic quality, but rather from their misinterpretation of long prompts and inability to comprehend complex semantic information.

Our proposed VisionDirector explicitly delegates semantic processing to a planner VLM \cite{yang2024visionzip}, which extracts goals, reasons about conflicts, and verifies each edit, while the diffusion backbones focus solely on rendering. 
Through this separation of concerns, the model preserves visual quality, generates detailed logs for debugging, and maintains a modular architecture. 
Moreover, a more capable editor or verifier can be substituted without retraining the planner.

\begin{table}[t]
  \centering
  \small
  \begin{tabular*}{\linewidth}{@{\extracolsep{\fill}} lcc}
    \toprule
    Method & Goal & $\geq 80$\% \\
    \midrule
    Baseline(Flux-Krea)        & 66.8 & 18.6 \\
    + Reprompting   & 69.0 & 22.7 \\
    + Best-of-N(N=4)     & 70.5 & 23.1 \\
    + Refinement    & 71.2 & 29.5 \\
    All Strategies  & 74.2 & 35.2 \\
    \bottomrule
  \end{tabular*}
  \vspace{-0.2cm}
  \caption{Ablation study on different optimization strategies (values in \%). ``Goal'' reports
  overall goal success rate, while $\geq$80\% indicates task-level high quality coverage. Each
  strategy is tested independently except the final row which combines all.}
  \label{tab:ablation}
  \vspace{-0.5cm}
\end{table}

\subsection{Broader Impact}

\mypara{Generate Diverse Training Dataset. }
Due to its high accuracy and fully automated nature, VisionDirector can be used to construct large-scale datasets, which can then be employed to train domain-specific or more advanced generative models.
Furthermore, the post-trained VisionDirector model can better adapt to the text distribution requirements of generation models. Compared with manually adjusting images through multiple prompt iterations, it can produce the desired images in fewer rounds without human intervention, significantly reducing token consumption and saving substantial time.

%% file: sec/6_related.tex
\section{Related Work}
\mypara{Diffusion generation and editing.}
Large diffusion models such as DDPM/Score models~\cite{ho2020denoising} and latent diffusion variants~\cite{rombach2021highresolution} have driven recent leaps in visual quality. Follow-up work tailors them for instruction-based editing, e.g., Instruct-Pix2Pix~\cite{brooks2023instructpix2pix} or the more recent unified pipelines (Step1X-Edit, BAGEL, UniWorld) summarized in Table~\ref{tab:imgedit}. Beyond images, diffusion-based video generation \cite{che2024gamegenxinteractiveopenworldgame} and editing \cite{yu2025positioninteractivegenerativevideo, yu2025surveyinteractivegenerativevideo} further magnify these limitations, as temporal consistency and long-horizon instruction adherence are difficult to maintain with single-shot prompting. These systems excel at local photorealism but still rely on single-shot prompts; none reason explicitly about dozens of heterogeneous goals. VisionDirector keeps these high-quality editors intact and adds a director layer that decomposes and monitors long instructions. 

\mypara{Benchmarking long-form instructions.}
Multi-goal evaluation has been explored in compositional prompts (GenEval~\cite{ghosh2023geneval}) and dense textual QA (DPG~\cite{hu2024ella}), but these suites cover at most two goals per prompt and focus on T2I only. ImgEdit~\cite{ye2025imgedit} and GEdit~\cite{liu2025step1x} evaluate instruction-based editing, yet their directives remain short and lack goal-level metadata. LGBench complements these efforts by providing dual modalities, 29k annotated goals, and a verifier that outputs structured pass/fail tags, enabling agent-style analysis.

%% file: sec/7_conclusion.tex
\section{Conclusion}
We presented LGBench, a dual-modality benchmark that stresses instruction following with 29k annotated goals, and VisionDirector, a director-style framework that uses VLMs to plan, verify, and refine diffusion outputs. The training-free controller already improves multi-goal fidelity by decomposing prompts, running micro-grid edits, and rolling back regressions; a subsequent GRPO phase further shortens trajectories and boosts quantitative scores across GenEval and ImgEdit. Our study suggests that pairing strong diffusion priors with lightweight reasoning agents is a practical path toward professional-grade creative tooling. Future work will scale the benchmark to video and 3D assets, tighten verifier alignment with human preference data, and explore collaborative editing loops that keep humans in the loop for subjective art direction.

%% file: main.bib
@String(ECCV= {Eur. Conf. Comput. Vis.})

@String(ECCV  = {ECCV})

@article{ghosh2023geneval,
  title={GenEval: An Object-Focused Framework for Evaluating Text-to-Image Alignment},
  author={Ghosh, Dhruba and Hajishirzi, Hannaneh and Schmidt, Ludwig},
  journal={Advances in Neural Information Processing Systems},
  volume={36},
  pages={52132--52152},
  year={2023}
}

@article{hu2024ella,
  title={ELLA: Equip Diffusion Models with LLM for Enhanced Semantic Alignment},
  author={Hu, Xiwei and Wang, Rui and Fang, Yixiao and Fu, Bin and Cheng, Pei and Yu, Gang},
  journal={arXiv preprint arXiv:2403.05135},
  year={2024}
}

@article{liu2025step1x,
  title={STEP1X-Edit: A Practical Framework for General Image Editing},
  author={Liu, Shiyu and Han, Yucheng and Xing, Peng and Yin, Fukun and Wang, Rui and Cheng, Wei and Liao, Jiaqi and Wang, Yingming and Fu, Honghao and Han, Chunrui and others},
  journal={arXiv preprint arXiv:2504.17761},
  year={2025}
}

@article{ye2025imgedit,
  title={ImgEdit: A Unified Image Editing Dataset and Benchmark},
  author={Ye, Yang and He, Xianyi and Li, Zongjian and Lin, Bin and Yuan, Shenghai and Yan, Zhiyuan and Hou, Bohan and Yuan, Li},
  journal={arXiv preprint arXiv:2505.20275},
  year={2025}
}

@article{shao2024deepseekmath,
  title={Deepseekmath: Pushing the limits of mathematical reasoning in open language models},
  author={Shao, Zhihong and Wang, Peiyi and Zhu, Qihao and Xu, Runxin and Song, Junxiao and Bi, Xiao and Zhang, Haowei and Zhang, Mingchuan and Li, YK and Wu, Yang and others},
  journal={arXiv preprint arXiv:2402.03300},
  year={2024}
}

@article{ho2020denoising,
  title={Denoising diffusion probabilistic models},
  author={Ho, Jonathan and Jain, Ajay and Abbeel, Pieter},
  journal={Advances in Neural Information Processing Systems},
  volume={33},
  pages={6840--6851},
  year={2020}
}

@article{rombach2021highresolution,
  title={High-resolution image synthesis with latent diffusion models},
  author={Rombach, Robin and Blattmann, Andreas and Lorenz, Dominik and Esser, Patrick and Ommer, Bj{\"o}rn},
  journal={IEEE Conference on Computer Vision and Pattern Recognition},
  year={2022}
}

@article{brooks2023instructpix2pix,
  title={InstructPix2Pix: Learning to follow image editing instructions},
  author={Brooks, Tim and Holynski, Aleksander and Efros, Alexei A},
  journal={IEEE Conference on Computer Vision and Pattern Recognition},
  year={2023}
}

@article{saharia2022photorealistic,
  title={Photorealistic text-to-image diffusion models with deep language understanding},
  author={Saharia, Chitwan and Chan, William and Saxena, Saurabh and Li, Lala and Whang, Jay and Denton, Emily L and Ghasemipour, Kamyar and Gontijo Lopes, Raphael and Karagol Ayan, Burcu and Salimans, Tim and others},
  journal={Advances in neural information processing systems},
  volume={35},
  pages={36479--36494},
  year={2022}
}

@article{wu2025qwen,
  title={Qwen-image technical report},
  author={Wu, Chenfei and Li, Jiahao and Zhou, Jingren and Lin, Junyang and Gao, Kaiyuan and Yan, Kun and Yin, Sheng-ming and Bai, Shuai and Xu, Xiao and Chen, Yilei and others},
  journal={arXiv preprint arXiv:2508.02324},
  year={2025}
}

@article{openai2024gpt4technicalreport,
  title={GPT-4 technical report},
  author={OpenAI},
  journal={arXiv preprint arXiv:2303.08774},
  year={2024}
}

@misc{flux1kreadev2025,
    author = {Sangwu Lee and Titus Ebbecke and Erwann Millon and Will Beddow
        and Le Zhuo and Iker Garcia-Ferrero and Liam Esparraguera
        and Mihai Petrescu and Gian Sass and Gabriel Menezes and Victor Perez},
    title = {FLUX.1 Krea [dev]},
    year = {2025},
    howpublished = {\url{https://github.com/krea-ai/flux-krea}},
}

@misc{flux2024,
    author={Black Forest Labs},
    title={FLUX},
    year={2024},
    howpublished={\url{https://github.com/black-forest-labs/flux}},
}

@article{batifol2025flux,
  title={FLUX. 1 Kontext: Flow Matching for In-Context Image Generation and Editing in Latent Space},
  author={Batifol, Stephen and Blattmann, Andreas and Boesel, Frederic and Consul, Saksham and Diagne, Cyril and Dockhorn, Tim and English, Jack and English, Zion and Esser, Patrick and Kulal, Sumith and others},
  journal={arXiv e-prints},
  pages={arXiv--2506},
  year={2025}
}

@article{xie2024show,
  title={Show-o: One single transformer to unify multimodal understanding and generation},
  author={Xie, Jinheng and Mao, Weijia and Bai, Zechen and Zhang, David Junhao and Wang, Weihao and Lin, Kevin Qinghong and Gu, Yuchao and Chen, Zhijie and Yang, Zhenheng and Shou, Mike Zheng},
  journal={arXiv preprint arXiv:2408.12528},
  year={2024}
}

@article{wang2024emu3,
  title={Emu3: Next-token prediction is all you need},
  author={Wang, Xinlong and Zhang, Xiaosong and Luo, Zhengxiong and Sun, Quan and Cui, Yufeng and Wang, Jinsheng and Zhang, Fan and Wang, Yueze and Li, Zhen and Yu, Qiying and others},
  journal={arXiv preprint arXiv:2409.18869},
  year={2024}
}

@misc{chen2023pixart,
      title={PixArt-$\alpha$: Fast Training of Diffusion Transformer for Photorealistic Text-to-Image Synthesis}, 
      author={Junsong Chen and Jincheng Yu and Chongjian Ge and Lewei Yao and Enze Xie and Yue Wu and Zhongdao Wang and James Kwok and Ping Luo and Huchuan Lu and Zhenguo Li},
      year={2023},
      eprint={2310.00426},
      archivePrefix={arXiv},
      primaryClass={cs.CV}
}

@inproceedings{esser2024scaling,
  title={Scaling rectified flow transformers for high-resolution image synthesis},
  author={Esser, Patrick and Kulal, Sumith and Blattmann, Andreas and Entezari, Rahim and M{\"u}ller, Jonas and Saini, Harry and Levi, Yam and Lorenz, Dominik and Sauer, Axel and Boesel, Frederic and others},
  booktitle={Forty-first international conference on machine learning},
  year={2024}
}

@inproceedings{ma2025janusflow,
  title={Janusflow: Harmonizing autoregression and rectified flow for unified multimodal understanding and generation},
  author={Ma, Yiyang and Liu, Xingchao and Chen, Xiaokang and Liu, Wen and Wu, Chengyue and Wu, Zhiyu and Pan, Zizheng and Xie, Zhenda and Zhang, Haowei and Yu, Xingkai and others},
  booktitle={Proceedings of the Computer Vision and Pattern Recognition Conference},
  pages={7739--7751},
  year={2025}
}

@article{chen2025janus,
  title={Janus-pro: Unified multimodal understanding and generation with data and model scaling},
  author={Chen, Xiaokang and Wu, Zhiyu and Liu, Xingchao and Pan, Zizheng and Liu, Wen and Xie, Zhenda and Yu, Xingkai and Ruan, Chong},
  journal={arXiv preprint arXiv:2501.17811},
  year={2025}
}

@article{qin2025lumina,
  title={Lumina-image 2.0: A unified and efficient image generative framework},
  author={Qin, Qi and Zhuo, Le and Xin, Yi and Du, Ruoyi and Li, Zhen and Fu, Bin and Lu, Yiting and Yuan, Jiakang and Li, Xinyue and Liu, Dongyang and others},
  journal={arXiv preprint arXiv:2503.21758},
  year={2025}
}

@article{cai2025hidream,
  title={HiDream-I1: A High-Efficient Image Generative Foundation Model with Sparse Diffusion Transformer},
  author={Cai, Qi and Chen, Jingwen and Chen, Yang and Li, Yehao and Long, Fuchen and Pan, Yingwei and Qiu, Zhaofan and Zhang, Yiheng and Gao, Fengbin and Xu, Peihan and others},
  journal={arXiv preprint arXiv:2505.22705},
  year={2025}
}

@article{gao2025seedream,
  title={Seedream 3.0 technical report},
  author={Gao, Yu and Gong, Lixue and Guo, Qiushan and Hou, Xiaoxia and Lai, Zhichao and Li, Fanshi and Li, Liang and Lian, Xiaochen and Liao, Chao and Liu, Liyang and others},
  journal={arXiv preprint arXiv:2504.11346},
  year={2025}
}

@article{zhang2023magicbrush,
  title={Magicbrush: A manually annotated dataset for instruction-guided image editing},
  author={Zhang, Kai and Mo, Lingbo and Chen, Wenhu and Sun, Huan and Su, Yu},
  journal={Advances in Neural Information Processing Systems},
  volume={36},
  pages={31428--31449},
  year={2023}
}

@article{jiang2025anyedit,
  title={Anyedit: Edit any knowledge encoded in language models},
  author={Jiang, Houcheng and Fang, Junfeng and Zhang, Ningyu and Ma, Guojun and Wan, Mingyang and Wang, Xiang and He, Xiangnan and Chua, Tat-seng},
  journal={arXiv preprint arXiv:2502.05628},
  year={2025}
}

@article{zhao2024ultraedit,
  title={Ultraedit: Instruction-based fine-grained image editing at scale},
  author={Zhao, Haozhe and Ma, Xiaojian Shawn and Chen, Liang and Si, Shuzheng and Wu, Rujie and An, Kaikai and Yu, Peiyu and Zhang, Minjia and Li, Qing and Chang, Baobao},
  journal={Advances in Neural Information Processing Systems},
  volume={37},
  pages={3058--3093},
  year={2024}
}

@inproceedings{xiao2025omnigen,
  title={Omnigen: Unified image generation},
  author={Xiao, Shitao and Wang, Yueze and Zhou, Junjie and Yuan, Huaying and Xing, Xingrun and Yan, Ruiran and Li, Chaofan and Wang, Shuting and Huang, Tiejun and Liu, Zheng},
  booktitle={Proceedings of the Computer Vision and Pattern Recognition Conference},
  pages={13294--13304},
  year={2025}
}

@article{zhang2025context,
  title={In-context edit: Enabling instructional image editing with in-context generation in large scale diffusion transformer},
  author={Zhang, Zechuan and Xie, Ji and Lu, Yu and Yang, Zongxin and Yang, Yi},
  journal={arXiv preprint arXiv:2504.20690},
  year={2025}
}

@article{lin2025uniworld,
  title={Uniworld: High-resolution semantic encoders for unified visual understanding and generation},
  author={Lin, Bin and Li, Zongjian and Cheng, Xinhua and Niu, Yuwei and Ye, Yang and He, Xianyi and Yuan, Shenghai and Yu, Wangbo and Wang, Shaodong and Ge, Yunyang and others},
  journal={arXiv preprint arXiv:2506.03147},
  year={2025}
}

@article{deng2505emerging,
  title={Emerging properties in unified multimodal pretraining, 2025},
  author={Deng, Chaorui and Zhu, Deyao and Li, Kunchang and Gou, Chenhui and Li, Feng and Wang, Zeyu and Zhong, Shu and Yu, Weihao and Nie, Xiaonan and Song, Ziang and others},
  journal={URL https://arxiv. org/abs/2505.14683}
}

@article{wu2025omnigen2,
  title={OmniGen2: Exploration to Advanced Multimodal Generation},
  author={Wu, Chenyuan and Zheng, Pengfei and Yan, Ruiran and Xiao, Shitao and Luo, Xin and Wang, Yueze and Li, Wanli and Jiang, Xiyan and Liu, Yexin and Zhou, Junjie and others},
  journal={arXiv preprint arXiv:2506.18871},
  year={2025}
}

@inproceedings{hu2023tifa,
  title={Tifa: Accurate and interpretable text-to-image faithfulness evaluation with question answering},
  author={Hu, Yushi and Liu, Benlin and Kasai, Jungo and Wang, Yizhong and Ostendorf, Mari and Krishna, Ranjay and Smith, Noah A},
  booktitle={Proceedings of the IEEE/CVF International Conference on Computer Vision},
  pages={20406--20417},
  year={2023}
}

@misc{esser2024scalingrectifiedflowtransformers,
      title={Scaling Rectified Flow Transformers for High-Resolution Image Synthesis}, 
      author={Patrick Esser and Sumith Kulal and Andreas Blattmann and Rahim Entezari and Jonas Müller and Harry Saini and Yam Levi and Dominik Lorenz and Axel Sauer and Frederic Boesel and Dustin Podell and Tim Dockhorn and Zion English and Kyle Lacey and Alex Goodwin and Yannik Marek and Robin Rombach},
      year={2024},
      eprint={2403.03206},
      archivePrefix={arXiv},
      primaryClass={cs.CV},
      url={https://arxiv.org/abs/2403.03206}, 
}

@article{yang2025visionthink,
  title={VisionThink: Smart and Efficient Vision Language Model via Reinforcement Learning},
  author={Yang, Senqiao and Li, Junyi and Lai, Xin and Yu, Bei and Zhao, Hengshuang and Jia, Jiaya},
  journal={arXiv preprint arXiv:2507.13348},
  year={2025}
}

@article{guo2025deepseek,
  title={Deepseek-r1 incentivizes reasoning in llms through reinforcement learning},
  author={Guo, Daya and Yang, Dejian and Zhang, Haowei and Song, Junxiao and Wang, Peiyi and Zhu, Qihao and Xu, Runxin and Zhang, Ruoyu and Ma, Shirong and Bi, Xiao and others},
  journal={Nature},
  volume={645},
  number={8081},
  pages={633--638},
  year={2025},
  publisher={Nature Publishing Group UK London}
}

@article{yang2024visionzip,
  title={VisionZip: Longer is Better but Not Necessary in Vision Language Models},
  author={Yang, Senqiao and Chen, Yukang and Tian, Zhuotao and Wang, Chengyao and Li, Jingyao and Yu, Bei and Jia, Jiaya},
  journal={arXiv preprint arXiv:2412.04467},
  year={2024}
}

@misc{bai2025qwen3vltechnicalreport,
      title={Qwen3-VL Technical Report}, 
      author={Shuai Bai and Yuxuan Cai and Ruizhe Chen and Keqin Chen and Xionghui Chen and Zesen Cheng and Lianghao Deng and Wei Ding and Chang Gao and Chunjiang Ge and Wenbin Ge and Zhifang Guo and Qidong Huang and Jie Huang and Fei Huang and Binyuan Hui and Shutong Jiang and Zhaohai Li and Mingsheng Li and Mei Li and Kaixin Li and Zicheng Lin and Junyang Lin and Xuejing Liu and Jiawei Liu and Chenglong Liu and Yang Liu and Dayiheng Liu and Shixuan Liu and Dunjie Lu and Ruilin Luo and Chenxu Lv and Rui Men and Lingchen Meng and Xuancheng Ren and Xingzhang Ren and Sibo Song and Yuchong Sun and Jun Tang and Jianhong Tu and Jianqiang Wan and Peng Wang and Pengfei Wang and Qiuyue Wang and Yuxuan Wang and Tianbao Xie and Yiheng Xu and Haiyang Xu and Jin Xu and Zhibo Yang and Mingkun Yang and Jianxin Yang and An Yang and Bowen Yu and Fei Zhang and Hang Zhang and Xi Zhang and Bo Zheng and Humen Zhong and Jingren Zhou and Fan Zhou and Jing Zhou and Yuanzhi Zhu and Ke Zhu},
      year={2025},
      eprint={2511.21631},
      archivePrefix={arXiv},
      primaryClass={cs.CV},
      url={https://arxiv.org/abs/2511.21631}, 
}

@misc{yu2025surveyinteractivegenerativevideo,
      title={A Survey of Interactive Generative Video}, 
      author={Jiwen Yu and Yiran Qin and Haoxuan Che and Quande Liu and Xintao Wang and Pengfei Wan and Di Zhang and Kun Gai and Hao Chen and Xihui Liu},
      year={2025},
      eprint={2504.21853},
      archivePrefix={arXiv},
      primaryClass={cs.CV},
      url={https://arxiv.org/abs/2504.21853}, 
}

@misc{yu2025positioninteractivegenerativevideo,
      title={Position: Interactive Generative Video as Next-Generation Game Engine}, 
      author={Jiwen Yu and Yiran Qin and Haoxuan Che and Quande Liu and Xintao Wang and Pengfei Wan and Di Zhang and Xihui Liu},
      year={2025},
      eprint={2503.17359},
      archivePrefix={arXiv},
      primaryClass={cs.CV},
      url={https://arxiv.org/abs/2503.17359}, 
}

@misc{che2024gamegenxinteractiveopenworldgame,
      title={GameGen-X: Interactive Open-world Game Video Generation}, 
      author={Haoxuan Che and Xuanhua He and Quande Liu and Cheng Jin and Hao Chen},
      year={2024},
      eprint={2411.00769},
      archivePrefix={arXiv},
      primaryClass={cs.CV},
      url={https://arxiv.org/abs/2411.00769}, 
}

@misc{saharia2022photorealistictexttoimagediffusionmodels,
      title={Photorealistic Text-to-Image Diffusion Models with Deep Language Understanding}, 
      author={Chitwan Saharia and William Chan and Saurabh Saxena and Lala Li and Jay Whang and Emily Denton and Seyed Kamyar Seyed Ghasemipour and Burcu Karagol Ayan and S. Sara Mahdavi and Rapha Gontijo Lopes and Tim Salimans and Jonathan Ho and David J Fleet and Mohammad Norouzi},
      year={2022},
      eprint={2205.11487},
      archivePrefix={arXiv},
      primaryClass={cs.CV},
      url={https://arxiv.org/abs/2205.11487}, 
}

@inproceedings{chu2024towards, 
  title={Towards Natural Language-Guided Drones: GeoText-1652 Benchmark with Spatial Relation Matching}, 
  author={Chu, Meng and Zheng, Zhedong and Ji, Wei and Wang, Tingyu and Chua, Tat-Seng}, 
  booktitle={ECCV}, 
  year={2024} 
}
